# A Transfer Learning-Based Approach to Marine Vessel Re-Identification

Guangmiao Zeng, Rongjie Wang, Wanneng Yu* and Anhui Lin

*Abstract*—**Marine vessel re-identification technology is an important component of intelligent shipping systems and an important part of the visual perception tasks required for marine surveillance. However, unlike the situation on land, the maritime environment is complex and variable with fewer samples, and it is more difficult to perform vessel re-identification at sea. Therefore, this paper proposes a transfer dynamic alignment algorithm and simulates the swaying situation of vessels at sea, using a well-camouflaged and similar warship as the test target to improve the recognition difficulty and thus cope with the impact caused by complex sea conditions, and discusses the effect of different types of vessels as transfer objects. The experimental results show that the improved algorithm improves the mean average accuracy (mAP) by 10.2% and the first hit rate (Rank1) by 4.9% on average.**

## I. INTRODUCTION

ACCORDING to the International Regulations for Collision Avoidance at Sea (COLREG) [1], vessels at sea weighing more than 300 tons and all passenger ships are required to be equipped with an automatic navigation system (AIS), which is designed to report their position, speed and course, but does not include their size, which is only identified as a point by the system. In addition, some small and medium-sized vessels such as fishing ships and small cargo ships are not mandated to install AIS systems, and even on large vessels AIS can be turned off voluntarily, which can affect the ship's judgment of the surrounding safety. Moreover, it is true that the size of surrounding obstacles can be judged by using shipboard radar, but the volume of the vessel varies greatly at different angles, and the radar identification is more difficult due to the wider view at sea and more distant vessel targets [2]. Nowadays, with the rapid development of computer vision technology, visible light cameras can play an important role in the detection of the marine environment around vessels.

The target re-identification method is an important branch of machine vision system, and the implementation of vessel re-identification method is beneficial to build a perfect marine situational awareness system [3], and this method has been more widely used on person [4-8] and vehicles [9-13], but compared to the stable shooting environment on land, the sea surface where the vessel is located shakes a lot and the target object tilts is very common, while in the existing data set of land environment, the target objects are almost smooth although they have different postures. Moreover, this problem is not considered in the current articles on vessel reidentification [14-16], which are all based on vessels that are moored in port or running smoothly as identification objects. The normal moving vessels will have different degrees of tilt because of the speed and sea condition, and will have different draught depths because of the different load, and the shape of the visible parts on the sea surface will also change. Therefore, the existing re-identification method is not fully applicable to the vessels in the heading.

In this paper, since the known vessel re-identification datasets VesselReid [14] and VesselReid-539 [15] are not open source, we created and labeled a vessel dataset based on the published resource acquisition method and simulated the sway occurring on the sea surface by rotating the vessel in a small angle. In addition, due to the difficulty of acquiring images of maritime targets and the fact that weather factors can have a large impact on recognition when sailing in complex sea conditions, different images of warships in multiple contexts and with a large degree of similarity were chosen for the test set. Then, we propose a dynamic alignment re-recognition method incorporating transfer learning in order to make it possible to train the network with the relevance features of the transferred images in a limited number of training images, thus improving the recognition accuracy of the network, which does not require secondary labeling of the local information of the images again. Finally, during the experiment, a variety of different classes of vessels are selected as transfer data, and the effects of different classes, different sample sizes, and whether or not the sea sway simulation method is used on the recognition accuracy of the network are discussed.

The rest of the paper is organized as follows. Section 2 presents related work, Section 3 details the dynamic alignment algorithm incorporating transfer learning and the creation of the loss function, Section 4 provides experimental analysis and discussion, and then concludes in Section 5.

This work was supported in part by the National Natural Science Foundation of China under Grant No. 52171308 and No. 51879118, in part by the Natural Science Foundation of Fujian Province No. 2020J01688, in part by the Key Projects of Science and Technology in Fujian Province No. 2021H0021, in part by the Science and Technology Support Project of Fujian Province No. B19101, in part by the Transportation Industry High-Level Technical Talent Training Project No.2019-014, in part by the Young Talent of Jimei University No. ZR2019006. *(Corresponding author: Wanneng Yu).*

Guangmiao Zeng, Rongjie Wang, Wanneng Yu and Anhui Lin are with the School of Marine Engineering, Jimei University, Xiamen 361021, China. (e-mail: gm.zeng@foxmail.com).

## II. RELATED WORKS

### A. Person Re-Identification

The most studied application of re-identification methods is the person re-identification problem, which mainly focuses on the improvement of the network structure so as to enhance the metric learning effect. Luo et al. proposed a local dynamic alignment method, which solves the problem of misalignment of human pose models without additional supervision and incorporates the global features for training [4]. Wang et al. improved the network structure by profiling the global average pooling layer and using an attention mechanism to improve the attention of spatial relations in elemental graphs [6]. Sun et al. improved the generalization effect for the metric model and proposed a semi-supervised consistent projection metric learning method to solve the biased estimation problem [7].

At the same time, there are still some studies from the perspective of datasets that do not use visible light cameras or add other dimensions of information to improve the recognition accuracy of existing algorithms, such as Cheng et al. used vehicle radar point cloud data as a dataset for person recognition and designed a corresponding re-identification network structure that not only protects the privacy of the person but also can be used in dim light environments [5]. Shao et al. extracted spatio-temporal information from real surveillance scenes and proposed a novel residual spatial attention module to improve the network performance by adding spatial-temporal constraints [8]. However, the vessel features and person features differ greatly, and the proposed algorithm based on person re-identification is not fully applicable to maritime targets.

### B. Vehicle Re-Identification

Vehicle re-identification is another direction that has been widely studied after person re-identification, and improvements to the network structure remain the main way to improve the performance of the algorithm. Martinel et al. added ring distillation blocks to the feature pyramid structure and used a new pooling layer to limit the effect of vehicle orientation while improving detection accuracy and detection speed [11]. Liu et al. incorporated group loss into the calculation of the loss function, which enhanced the network's learning of the connection between global and local features of the vehicle and improved the detail recognition of similar samples [12]. Xiong et al. built a filter matrix in the network by incorporating an intelligent driver model for estimating the arrival time of upstream vehicles, thus reducing the influence of the environment on vehicle recognition [13].

Not only that, some studies have supplemented different observation perspectives with new features based on the characteristics of the vehicle structure itself, such as Hao et al. explored the effect of vehicle pose on the effect of re-identification and proposed a pose robust feature for solving the pose obstacle problem, which improved the recognition capability of the network in multiple datasets [10]. Since supervised learning methods need to pre-label the dataset, which is relatively time-consuming and labor-intensive, Zheng et al. proposed a viewpoint recognition progressive clustering framework for unsupervised vehicle re-identification, which first divides the entire feature space into different subspaces based on a predictive viewpoint, and then performs progressive clustering to mine the exact relationships between samples [9]. Since vessels are at sea, their working environment differs greatly from that of land where vehicles are located, and the idea of using methods similar to unsupervised learning to achieve time and effort savings is desirable because few methods can achieve accurate acquisition of local features without the use of supervision.

### C. Vessel Re-Identification

There have been many vessel detections in recent years, but most of them focus on remote sensing detection using synthetic aperture radar [17-19]. However, for the re-identification problem, remote sensing information does not provide detailed features well, and the requirement of real-time detection cannot be achieved due to the top view taken by satellite. Therefore, it is necessary to use vessel pictures taken directly on shore or on board as a dataset. Ghahremani et al. proposed an identity-oriented re-identification network with an improved triadic loss function to build the VesselReid dataset [14]. Qiao et al. not only built a large ship dataset, VesselReid-539, but also locally labeled the ship into four aspects: stem, stern, side freeboard and superstructure, and trained it using quintuple loss, which led to a better recognition [15]. After that, vessel tracking was achieved by a re-recognition method using a gated cyclic cell model incorporating an attention mechanism to predict the maneuver probability of the vessel [16].

However, although the above two methods achieve vessel re-identification relatively well, the vessels in the data set are basically in better conditions, driving smoothly without swaying. But in practice, the sea weather is complicated and changeable, and the wind and waves are not calm, so it is common for the vessel to have swaying, especially for the small size of the vessel. Moreover, due to the far view at sea, and the foggy sky and sea reflection effect, the color of the same vessel under different viewpoints varies greatly, and the high-definition pictures in the dataset do not accurately reflect the actual effect. Therefore, a targeted study is needed to investigate the influence of vessel sway conditions and faded colors on ship recognition due to the sea surface. The proposed method differs from the existing vessel re-recognition methods in that it does not need to train images of multiple views of the same vessel at the same time, reduces the preliminary work of data set processing, increases the generalization ability of the algorithm, and improves the re-identification ability of vessels in complex sea conditions by using the sea sway simulation method for learning.

### D. Vessel Re-Identification

The core of transfer learning is to find the similarity between the source and target domains. For the target of warships, where the data set is not abundant, it becomes difficult to use limited data to improve the test accuracy, and the transfer learning method can use multiple types of vessels as the source domain data to assist the network in learning the target domain warships.

Among them are methods based on statistical feature transformations, and after the classical transfer component analysis method was proposed [20], the maximum mean difference (MMD) distance is widely used to compute the difference in edge distribution between the source and target domains, such as dynamic distribution adaptation [21], transfer methods combined with deep neural networks [22-23], and deep dynamic transfer methods [24]. Besides, geometric features can be used for similarity correlation, such as the classical subspace alignment method [25], as well as first-order feature alignment [26] and second-order feature alignment [27] for the source and target domains for alignment of feature distributions, and neural networks can also calculate the second-order feature distances of the source and target domains as loss functions for optimization.

Therefore, transfer learning can be accomplished by decreasing the statistical feature distance and geometric feature distance between warships and other classes of vessels in order to increase the similarity between them as the vessel class.

## III. Methodology

Dynamic Alignment Algorithm (AlignedReID) [4] is a detection method proposed by Megvii Inc. in 2018 for person re-identification, which uses the similarity of the same parts of a person for dynamic planning. In this paper, local similarity is fused with transfer learning method to propose a re-identification method for vessel targets, Transfer Dynamic Alignment Algorithm (Tran-AlignedReID), and the struture is shown in Figure 1.

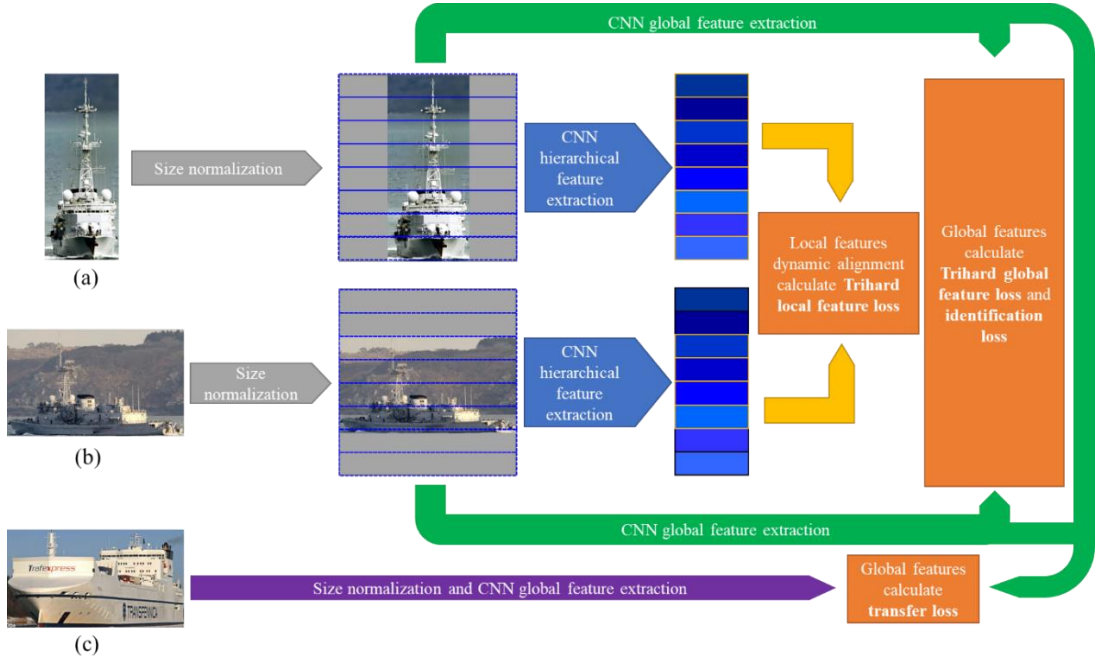

**Fig. 1.** The structure of Tran-AlignedReID.

Due to the large difference between the vessel target and person target contours, and the overall aspect ratio of the target is not consistent when the same vessel is observed from different angles, as shown in the front view a and side view b in Figure 1. Therefore, it is necessary to first perform a size normalization operation to make the target aspect ratio consistent by using the filling method of adding gray bars, and then scaling to the same resolution as the input data for the neural network. Then the input data of multiple images are horizontally segmented and divided equally into multiple rectangular regions of the same size and matched using the similarity relationship to calculate the local loss. The cross entropy loss of global features is also calculated with the input data as a whole. And the identification loss is calculated based on the label data of each group of images. After that, transfer learning is performed using other types of vessels to calculate the transfer loss. Finally, the results of these four loss functions are combined and the network weights are updated.

### A. Local Feature Analysis

For each ship target map with different shapes, it will be used as the input data of the convolutional neural network in a uniform format and feature extraction will be performed with a backbone network structure of ResNet50 [2], which is compressed for dynamic feature alignment, and the calculation procedure is shown in Figure 2.

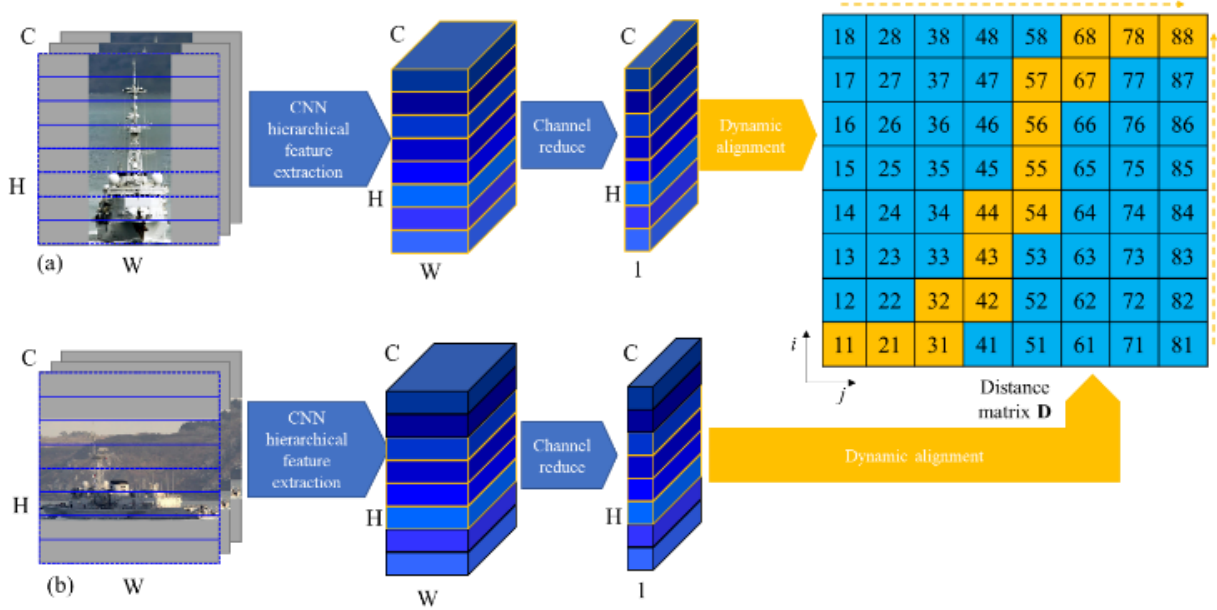

**Fig. 2.** Dynamic alignment method for local features.

First, the features are extracted from the input data in three dimensions: number of channels, height, and width (C×H×W), and divided into $r$ equal parts in the vertical direction, with $r$ equal to 8 in Figure 2. After that, the extracted features need to be further compressed in the W direction to form a compressed feature matrix of size C×H×1. Before dynamic alignment, the compressed feature matrices $l_a$ and $l_b$ of Image (a) and Image (b) need to be normalized to [0,1) for computational convenience, and the similarity distances $d_{i,j}$ of the compressed features in the W direction of the two images need to be calculated, as shown in (1).

$$d_{i,j} = \frac{e^{\left\| l_a^i - l_b^j \right\|_2} - 1}{e^{\left\| l_a^i - l_b^j \right\|_2} + 1} \quad i, j \in 1, 2, 3, ..., r,$$  (1)

where $l_a^i$ and $l_b^j$ are the feature vectors of layer $i$ and layer $j$ in the compressed feature matrix of Image a and Image b.

In order to obtain the effective features more accurately, the minimum value of the sum of similarity distances needs to be calculated by the distance matrix **D**. Thus, the local feature matching method with the highest similarity between the two images is found, as shown in (2) to (3).

$$D_{i,j} = \begin{cases} d_{i,j} & i=1, j=1 \\ D_{i-1,j} + d_{i,j} & i>1, j=1 \\ D_{i,j-1} + d_{i,j} & i=1, j>1 \\ \min\left(D_{i-1,j}, D_{i,j-1}\right) + d_{i,j} & i>1, j>1 \end{cases},$$  (2)

$$D_l\left(\mathrm{a,b}\right) = D_{r,r},$$  (3)

where is illustrated by the distance matrix **D** in Figure 2. $D_{i,j}$ is the sum of the minimum local distances from $d_{1,1}$ to $d_{i,j}$. $D_l(\mathrm{a,b})$ is the minimum total local distance between Image (a) and Image (b). The calculation procedure is shown by the golden route in **D**. Its value can be represented by the value in the upper right cell of **D**.

### B. Global Feature Analysis

For feature extraction of images, not only the local features need to be mined, but also the global features cannot be ignored. The network has been used to extract features of different depths from the input data, and is able to calculate the global feature distance and identification loss between samples, and the calculation process is shown in Figure 3.

For each image in each set of $N$ input data, the feature matrix (C×H×W) is first extracted using the same method as in the local feature analysis. After that, the whole is mainly transformed into the feature vector ($C'$×1×1). Also taking Image (a) and Image (b) as an example, the global feature distance $d_g(\mathrm{a,b})$ is calculated as shown in (4).

$$d_g(\mathrm{a,b}) = \left\| f_a - f_b \right\|_2,$$  (4)

where $f_a$ and $f_b$ denote the global feature vectors of Image (a) and Image (b), respectively.

In addition, by further compressing the $C'$ channels in the feature vector ($C' \times 1 \times 1$) into $M$ channels (ID ≤ N), the same vessel from different angles in each group of $N$ samples is noted as the same class, and $M$ is the number of feature classes in each group of samples. From this, the recognition loss $L_{\mathrm{ID}}$ of each group of samples is calculated as shown in (5) to (6).

$$L_{\mathrm{ID}} = \frac{1}{N}\sum_{n=1}^{N} L_n = -\frac{1}{N}\sum_{n=1}^{N}\sum_{m=1}^{M} q_{nm} \log\left(p_{nm}\right),$$  (5)

$$q_{nm} = \begin{cases} 1-\varepsilon & n=m \\ \dfrac{\varepsilon}{M-1} & n \neq m \end{cases},$$  (6)

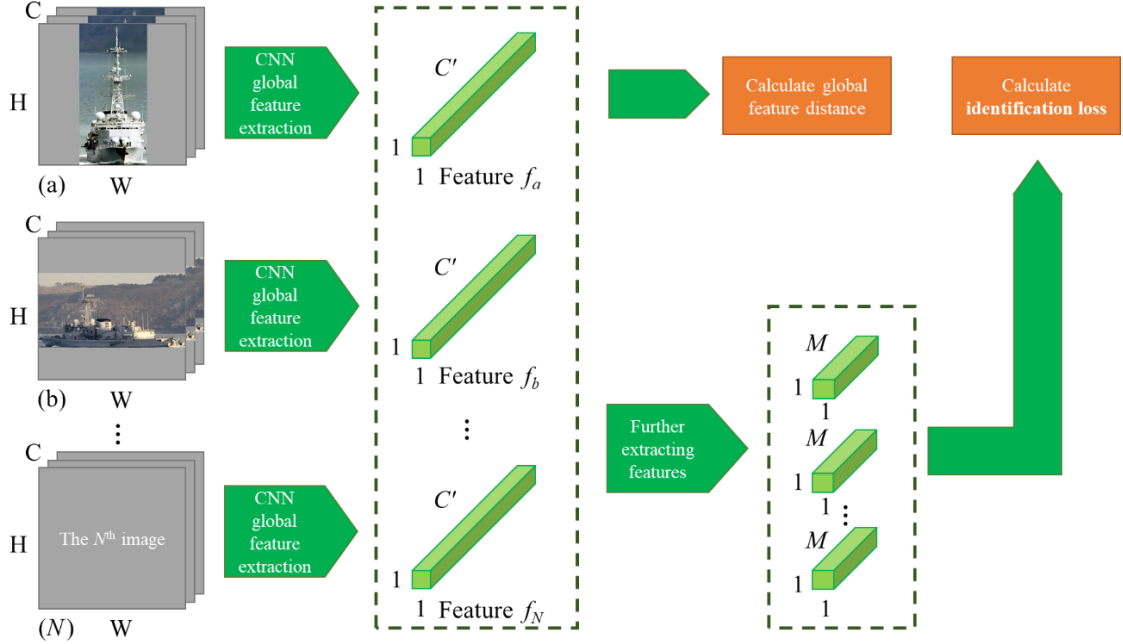

**Fig. 3.** The calculation procedure of global feature distance and identification loss.

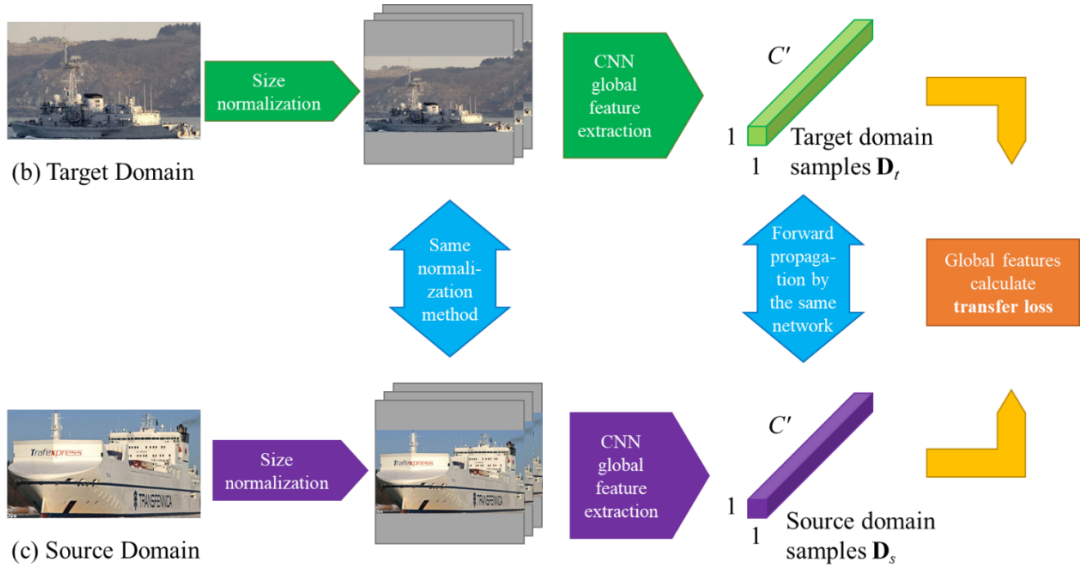

**Fig. 4.** The procedure for calculating transfer loss.

where $M$ is the total number of feature classes, $m$ denotes the $m^{th}$ feature number of a feature class, $n$ denotes the $n^{th}$ feature number of a sample, $p_{nm}$ is the predicted probability value of $n$ belonging to class $m$, and $q_{nm}$ is the soft label sign function after label smoothing operation. The overfitting phenomenon caused by too absolute weights is improved by changing the hard target to a soft target.

### C. Transfer Loss

Since the observation of maritime targets is different from that of land targets, the target and observation equipment are both at sea, and irregular shaking caused by the influence of waves will make the recognition more difficult. Therefore, there is a need to extract features from maritime targets in a variety of ways so that sample features can be fully utilized and recognition accuracy can be improved.

Using the method based on statistical feature transformation and geometric feature transformation, transfer learning is performed on multiple types of ship images by studying the similarity of the domains, so as to calculate the transfer loss, and the procedure is shown in Figure 4.

First, the source domain image S and the target domain image T are put into the same convolutional network by the same normalization. After the global feature extraction, the kernel trick is applied to calculate the value of MMD as shown in (7) to (9).

$$k(s,t) = \phi(s) \cdot \phi(t) \quad \phi(x) : \mathcal{X} \to \mathcal{H}, \qquad (7)$$

$$\text{MMD}(S,T)$$

$$= \left\| \frac{1}{N_s} \sum_{\alpha=1}^{N_s} \phi(\mathbf{x}_\alpha) - \frac{1}{N_t} \sum_{\beta=1}^{N_t} \phi(\mathbf{x}_\beta) \right\|_{\mathcal{H}}^2$$

$$= \text{tr}\left( \begin{bmatrix} \phi(\mathbf{x}_s) & \phi(\mathbf{x}_t) \end{bmatrix} \begin{bmatrix} \frac{1}{N_s^2}\mathbf{1}\mathbf{1}^{\mathbf{T}} & \frac{-1}{N_s N_t}\mathbf{1}\mathbf{1}^{\mathbf{T}} \\ \frac{-1}{N_s N_t}\mathbf{1}\mathbf{1}^{\mathbf{T}} & \frac{1}{N_t^2}\mathbf{1}\mathbf{1}^{\mathbf{T}} \end{bmatrix} \begin{bmatrix} \phi(\mathbf{x}_s) \\ \phi(\mathbf{x}_t) \end{bmatrix} \right)$$

$$= \text{tr}\left( \begin{bmatrix} \phi(\mathbf{x}_s) \\ \phi(\mathbf{x}_t) \end{bmatrix} \begin{bmatrix} \phi(\mathbf{x}_s) & \phi(\mathbf{x}_t) \end{bmatrix} \begin{bmatrix} \frac{1}{N_s^2}\mathbf{1}\mathbf{1}^{\mathbf{T}} & \frac{-1}{N_s N_t}\mathbf{1}\mathbf{1}^{\mathbf{T}} \\ \frac{-1}{N_s N_t}\mathbf{1}\mathbf{1}^{\mathbf{T}} & \frac{1}{N_t^2}\mathbf{1}\mathbf{1}^{\mathbf{T}} \end{bmatrix} \right),$$

$$= \text{tr}\left( \begin{bmatrix} \langle \phi(\mathbf{x}_s), \phi(\mathbf{x}_s) \rangle & \langle \phi(\mathbf{x}_s), \phi(\mathbf{x}_t) \rangle \\ \langle \phi(\mathbf{x}_t), \phi(\mathbf{x}_s) \rangle & \langle \phi(\mathbf{x}_t), \phi(\mathbf{x}_t) \rangle \end{bmatrix} \mathbf{U} \right)$$

$$= \text{tr}\left( \begin{bmatrix} K_{s,s} & K_{s,t} \\ K_{t,s} & K_{t,t} \end{bmatrix} \mathbf{U} \right)$$

$$(8)$$

$$(U)_{\alpha\beta} = \begin{cases} \frac{1}{N_s N_s} & \mathbf{x}_\alpha, \mathbf{x}_\beta \in \mathbf{D}_s \\ \frac{1}{N_t N_t} & \mathbf{x}_\alpha, \mathbf{x}_\beta \in \mathbf{D}_t \\ \frac{-1}{N_s N_t} & \text{otherwise} \end{cases}, \qquad (9)$$

where $\phi(x)$ is the mapping function from the input space $\mathcal{X}$ to the feature space $\mathcal{H}$, $k(s,t)$ is the kernel function, which is defined as the inner product of the mapping. And $<\cdot,\cdot>$ denotes the inner product operation, $\text{tr}(\cdot)$ denotes the trace of the matrix, $\mathbf{D}_s$ and $\mathbf{D}_t$ denote the samples in the source and target domains, respectively. $\mathbf{x}$ is the eigenvector in the sample, $(U)_{\alpha\beta}$ denotes the value of the element in the $\alpha^{\text{th}}$ row and $\beta^{\text{th}}$ column of the matrix $\mathbf{U}$, and $N_s$ and $N_t$ are the total number of samples in each group of source and target domains, respectively.

In this paper $N_s$ is equal to $N_t$ and the radial basis function $k(s,t)$ is used as the kernel function to calculate the Euclidean distance of the two vectors as shown in (10).

$$k(s,t) = e^{\frac{-\|s-t\|^2}{2\sigma^2}}, \qquad (10)$$

where $\sigma$ is the bandwidth and controls the radial range of action of the function.

Then, based on the data after the global feature extraction, the second-order feature alignment is performed on the source and target domain samples using the subspace transformation method to calculate their second-order statistical feature distances (CORAL), as shown in (11).

$$\text{CORAL}(S,T) = \frac{1}{4C'^2} \|\mathbf{Cov}_s - \mathbf{Cov}_t\|_F^2, \qquad (11)$$

where $\mathbf{Cov}_s$ and $\mathbf{Cov}_t$ denote the characteristic covariance matrices of the source domain samples and the target domain samples.

$$\mathbf{Cov}_s = \frac{1}{N_s-1}\left( \mathbf{D}_s^{\mathbf{T}}\mathbf{D}_s - \frac{1}{N_s}\left(\mathbf{1}^{\mathbf{T}}\mathbf{D}_s\right)^{\mathbf{T}}\left(\mathbf{1}^{\mathbf{T}}\mathbf{D}_s\right) \right),$$

$$\mathbf{Cov}_t = \frac{1}{N_t-1}\left( \mathbf{D}_t^{\mathbf{T}}\mathbf{D}_t - \frac{1}{N_t}\left(\mathbf{1}^{\mathbf{T}}\mathbf{D}_t\right)^{\mathbf{T}}\left(\mathbf{1}^{\mathbf{T}}\mathbf{D}_t\right) \right).$$

Thus, the combined transfer loss function $L_{\text{tran}}$ can be calculated from (12) as follows.

$$L_{\text{tran}}(S,T) = \gamma \text{MMD}(S,T) + \rho \text{CORAL}(S,T), \qquad (12)$$

where $\gamma$ and $\rho$ are the proportional weights, which are set to 1 and 0.001 in this paper, in order to make the two distance losses of the same order of magnitude.

### D. Joint Loss Function

In order to make better use of local feature distance and global feature distance so that the network can learn better representations, it is necessary to perform similarity calculation for each image in a set of images with other images after feature extraction. And according to the calculation results, the positive sample pair $d^{a,u}$ with the lowest similarity and the negative sample pair $d^{a,v}$ with the highest similarity are picked out for each image. Thus calculating its hard-sample sampling triplet loss (TriHard Loss) $L_{\text{tri}}$ based on local features and global features. Using Image (a) as an example, the calculation is shown in (13).

$$L_{\text{tri}} = \frac{1}{P \times Q} \sum_{a \in batch} \left( \max_{u \in A} D_{\text{tri}}^{a,u} - \min_{v \in B} D_{\text{tri}}^{a,v} + \eta \right)_+, \qquad (13)$$

where $D_{\text{tri}}$ is the total distance of features between two images, $\lambda$ is the hyperparameter to balance the global distance and local distance, which is set to 1 in this paper.

$$D_{\text{tri}} = D_l + \lambda d_g.$$

Therefore, by analyzing the local features and global features between images and introducing the transfer learning metric, the joint feature loss function $Loss$ can be calculated from (14).

$$Loss = L_{\text{tri}} + L_{\text{ID}} + L_{\text{tran}}. \qquad (14)$$

## IV. Experimental Results and Analysis

### A. Dataset

The dataset images for this paper are taken from the website Marine Traffic published in the maritime ship image dataset VesselID-539 (http://www.marinetraffic.com). The images were taken by videographers around the world on board vessels or on shore at different locations and times, and the same vessel presents a rich variety of poses due to the different times and angles from which the images were taken.

Among different types of ships, warships have high standards for concealment and flexibility, leading to a more homogeneous color, relatively fast speed, and flexible attitude, which have higher navigational demands compared to civilian vessels. Therefore, this paper will use warship pictures as training set and test set, and other types of vessels as transfer dataset for experiments.

There are 163 vessels and 4780 images in the vessel image dataset collected in this paper, and the specific categories and numbers are shown in Table I. And the images of warship classification are divided into training set and test set in the ratio of 1:1, and the query set accounts for 20% in the test set.

TABLE I
TYPE AND NUMBER OF VESSELS IN THE DATASET

| Vessel type | IDs | Total Images |
|---|---|---|
| Warship | 24 | 394 |
| Passenger ship | 23 | 692 |
| Sailboat | 22 | 870 |
| High speed ship | 21 | 641 |
| Cargo ship | 22 | 596 |
| Tug | 6 | 196 |
| Tanker | 23 | 677 |
| Fishing boat | 22 | 714 |

Since in practice, the surveillance camera is loaded on the vessel, it will undergo different degrees of multi-degree of freedom rocking in different sea conditions. Therefore, the data set is rotated randomly from -10° to 10° with the center of the image as the center point so as to simulate the sea sway (SSS), and the rotation process will cause part of the vessel's hull in the image to overflow out of the rectangular frame, and the overflowed part is cropped to simulate the situation that the visible part of the vessel's hull above the sea surface is different under different draught conditions.

Moreover, when the sea conditions are rough and windy, the vessel itself will be obscured by the waves, which leads to further reduction of the overall characteristics of the vessel.

Therefore, using the SSS method can improve the recognition ability of the network in the case of vessel tilting and obscuration.

### B. Experimental Procedure and Results

The algorithms in this paper were implemented on the open source neural network framework Pytorch (3.7.10). The computational workstation configuration contains 1 GPU (GeForce RTX 3090), CPU (AMD Ryzen 9 3950x 16 Core/ 3.5 GHz/72 M), and 128 G RAM. The dataset input images have an aspect ratio of 512×1024.

With the original AlignedReid algorithm, the warship images can be trained and other types of vessel images are used to supplement the training set. After training, the mean average precision (mAP) and the first hit rate (Rank1) of the network are shown in Table II.

As can be seen in Table II, the recognition accuracy of the network trained by the SSS images is 9.4% higher on average in mAP accuracy and 8.5% higher on average in Rank1 accuracy than that trained by the non-SSS images because the test set was subjected to SSS. Moreover, compared with training using only warships, using other types of vessel images as a supplement to the training set improved their mAP accuracy by 5.7% and 4.4% on average and Rank1 accuracy by 6.9% and 2.1% on average in the non-SSS case and the SSS case. Therefore, if the vessel is swayed due to waves, it is not sufficient to train only with the smooth images in the published dataset, and the training needs to enhance the data by simulating the sea sway on the images.

TABLE II
RESULTS OF VESSEL RE-IDENTIFICATION WITH ALIGNEDREID

| Training set | SSS or not | mAP (%) | Rank1 (%) | Rank5 (%) | Rank10 (%) | Rank20 (%) |
|---|---|---|---|---|---|---|
| Warships only | No | 43.8 | 82.9 | 94.3 | 97.1 | 97.1 |
| | Yes | 53.2 | 91.4 | 94.3 | 94.3 | 100.0 |
| +Passenger ship | | 50.3 | 91.4 | 91.4 | 94.3 | 97.1 |
| +Sailboat | | 43.0 | 88.6 | 91.4 | 100.0 | 100.0 |
| +High speed ship | | 54.7 | 94.3 | 94.3 | 97.1 | 100.0 |
| +Cargo ship | No | 48.1 | 88.6 | 91.4 | 91.4 | 100.0 |
| +Tug | | 52.8 | 88.6 | 91.4 | 97.1 | 100.0 |
| +Tanker | | 51.2 | 91.4 | 97.1 | 97.1 | 100.0 |
| +Fishing boat | | 46.2 | 85.7 | 91.4 | 94.3 | 100.0 |
| +Passenger ship | | 56.8 | 94.3 | 97.1 | 97.1 | 100.0 |
| +Sailboat | | 59.0 | 94.3 | 97.1 | 100.0 | 100.0 |
| +High speed ship | | 59.9 | 91.4 | 97.1 | 100.0 | 100.0 |
| +Cargo ship | Yes | 52.5 | 94.3 | 94.3 | 97.1 | 100.0 |
| +Tug | | 55.7 | 88.6 | 94.3 | 100.0 | 100.0 |
| +Tanker | | 58.0 | 97.1 | 97.1 | 100.0 | 100.0 |
| +Fishing boat | | 61.5 | 94.3 | 97.1 | 100.0 | 100.0 |

TABLE III

RESULTS OF VESSEL RE-IDENTIFICATION WITH TRAN-ALIGNEDREID
(TRAINING SET WITHOUT SSS)

| Transferring set | SSS or not | mAP (%) | Rank1 (%) | Rank5 (%) | Rank10 (%) | Rank20 (%) |
|---|---|---|---|---|---|---|
| / | No | 43.8 | 82.9 | 94.3 | 97.1 | 97.1 |
| Passenger ship | | 55.2 | 97.1 | 97.1 | 97.1 | 97.1 |
| Sailboat | | 50.0 | 91.4 | 97.1 | 97.1 | 100.0 |
| High speed ship | | 55.1 | 94.3 | 94.3 | 97.1 | 100.0 |
| Cargo ship | No | 58.9 | 91.4 | 94.3 | 100.0 | 100.0 |
| Tug | | 57.6 | 97.1 | 97.1 | 97.1 | 100.0 |
| Tanker | | 51.8 | 91.4 | 97.1 | 97.1 | 97.1 |
| Fishing boat | | 52.9 | 94.3 | 97.1 | 97.1 | 97.1 |
| Passenger ship | | 53.2 | 94.3 | 100.0 | 100.0 | 100.0 |
| Sailboat | | 49.6 | 94.3 | 94.3 | 100.0 | 100.0 |
| High speed ship | | 57.0 | 94.3 | 100.0 | 100.0 | 100.0 |
| Cargo ship | Yes | 58.2 | 100.0 | 100.0 | 100.0 | 100.0 |
| Tug | | 53.6 | 97.1 | 97.1 | 100.0 | 100.0 |
| Tanker | | 54.1 | 97.1 | 100.0 | 100.0 | 100.0 |
| Fishing boat | | 60.7 | 94.3 | 97.1 | 97.1 | 97.1 |

TABLE IV
RESULTS OF VESSEL RE-IDENTIFICATION WITH TRAN-ALIGNEDREID
(TRAINING SET WITH SSS)

| Transferring set | SSS or not | mAP (%) | Rank1 (%) | Rank5 (%) | Rank10 (%) | Rank20 (%) |
|---|---|---|---|---|---|---|
| / | Yes | 53.2 | 91.4 | 94.3 | 94.3 | 100.0 |
| Passenger ship | | 58.8 | 91.4 | 97.1 | 97.1 | 100.0 |
| Sailboat | | 62.1 | 91.4 | 91.4 | 97.1 | 100.0 |
| High speed ship | | 58.9 | 91.4 | 94.3 | 97.1 | 100.0 |
| Cargo ship | No | 63.8 | 94.3 | 94.3 | 97.1 | 100.0 |
| Tug | | 60.9 | 97.1 | 97.1 | 100.0 | 100.0 |
| Tanker | | 55.9 | 94.3 | 97.1 | 97.1 | 97.1 |
| Fishing boat | | 65.1 | 97.1 | 100.0 | 100.0 | 100.0 |
| Passenger ship | | 62.5 | 94.3 | 97.1 | 100.0 | 100.0 |
| Sailboat | | 63.5 | 100.0 | 100.0 | 100.0 | 100.0 |
| High speed ship | | 65.4 | 97.1 | 97.1 | 100.0 | 100.0 |
| Cargo ship | Yes | 63.3 | 94.3 | 94.3 | 100.0 | 100.0 |
| Tug | | 63.5 | 94.3 | 97.1 | 100.0 | 100.0 |
| Tanker | | 61.6 | 97.1 | 100.0 | 100.0 | 100.0 |
| Fishing boat | | 64.2 | 97.1 | 97.1 | 100.0 | 100.0 |

It can also be seen in Table 2 that even if all other types of vessels are used as supplementary training sets for warships, the improvement in their test accuracy is relatively limited, and for the convolutional neural network whose training set is a warship, all that the other types of vessels can provide is as global features of the vessel, and the network does not need to know too many details about it. Therefore, the study of domain similarity can be well enhanced by transfer learning methods.

With the training set without SSS, other classes of ships are trained by the transfer learning approach and tested using the test set after SSS, and the recognition results are shown in Table III.

As can be seen from Table III, compared to adding other classes of vessel images as expanded data to the training set, fusing them with the original algorithm in a transfer learning method resulted in 5.0% improvement in the average test accuracy on mAP and 4.1% improvement in Rank1 accuracy. And if the images are simulated with SSS and then used as transfer data, the mAP accuracy and Rank1 accuracy can be improved by another 0.7% and 2.1% compared to those without SSS.

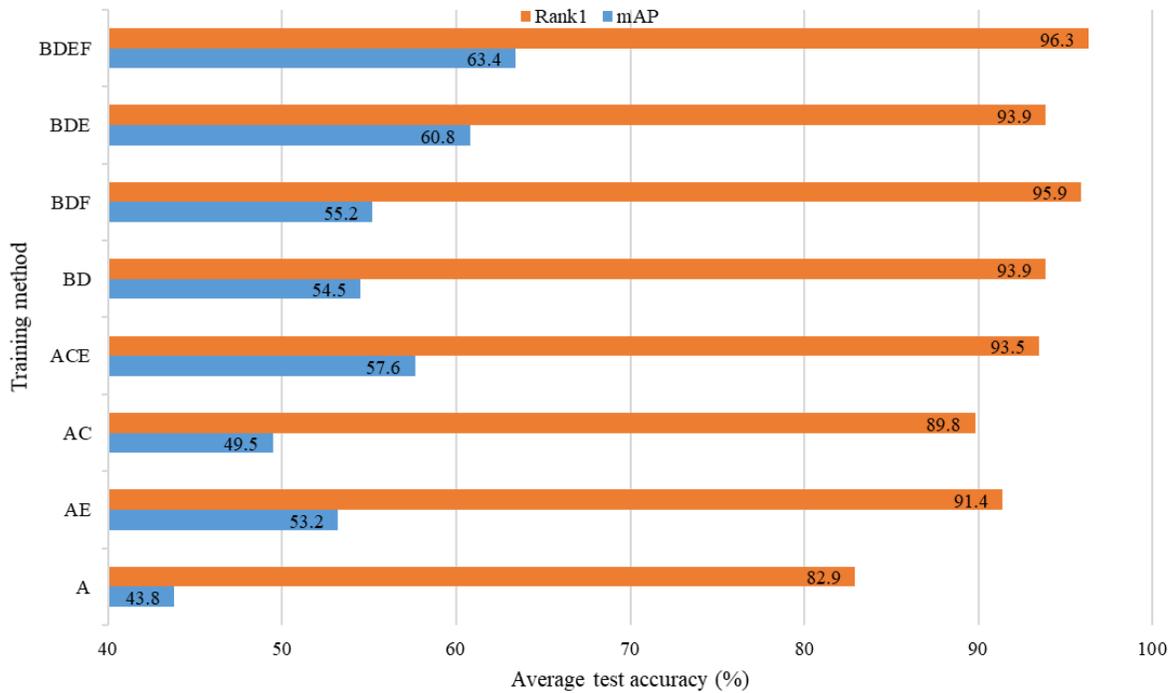

**Fig. 5.** The average test accuracy of the network before and after the improvement with different training methods, A:AlignedReid; B: Tran- AlignedReID; C: Expanded training set; D: Transferring dataset; E: Training set with SSS; F: Transferring dataset with SSS.

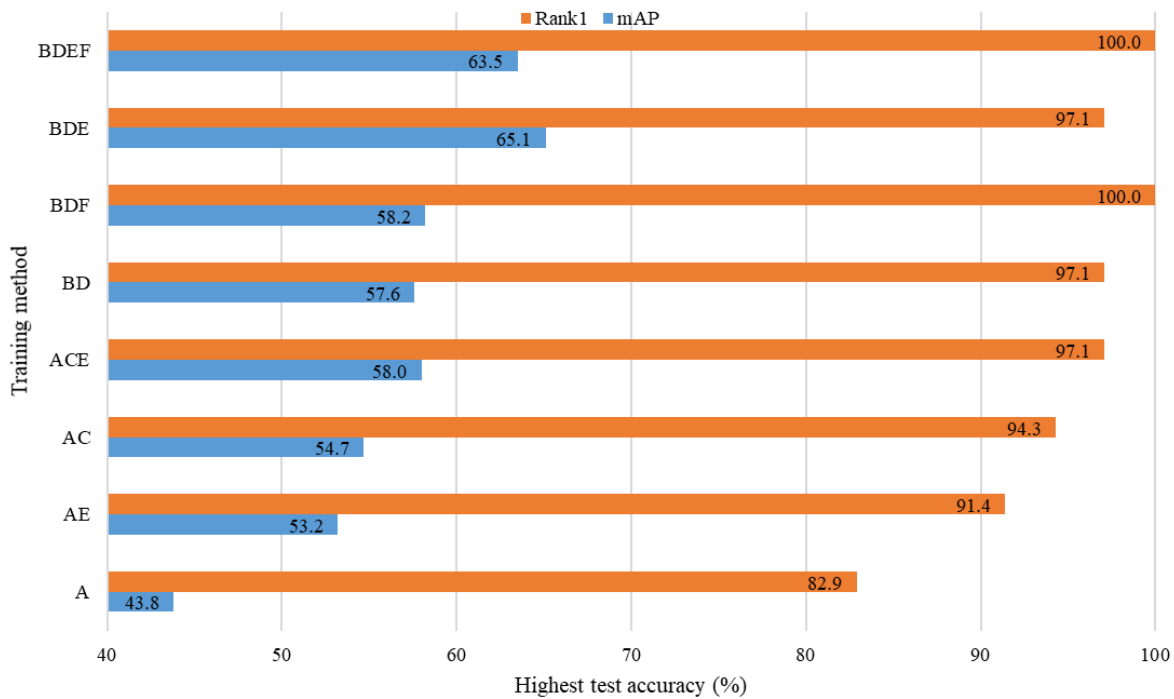

**Fig. 6.** The highest test accuracy of the network before and after the improvement with different training methods, A:AlignedReid; B: Tran- AlignedReID; C: Expanded training set; D: Transferring dataset; E: Training set with SSS; F: Transferring dataset with SSS.

Similarly, after subjecting the training set to SSS, the transfer fusion training is performed again and the obtained recognition results are shown in Table IV. It can be seen that the average test accuracy is improved by 3.2% in mAP and 0.4% in Rank1 accuracy even if the vessel images without SSS are used as transfer data, compared with the images with SSS directly as the expanded training set. The mAP and Rank1 accuracies improved by another 2.6% and 2.5% when the vessel images with SSS are used as transfer data compared to those without SSS.

The improved network before and after was trained separately for different classes of vessels, and its average test accuracy under different training methods is shown in Figure 5, and the highest test accuracy is shown in Figure 6.
From Fig. 5 and Fig. 6, it can be seen that both the average and the highest test accuracies, the results of fusing transfer learning methods for training are better than expanding the images directly to the training set.

Moreover, for the test images that undergo SSS, whether the training set undergoes SSS has a greater impact on the test accuracy, and the improvement in accuracy from transferring the dataset after SSS is relatively small. However, it also better illustrates that the transferred images do not emphasize the matching of their local features with the training set images, but rather the similarity association between the global features.

The highest test accuracies listed in Figure 6 are with Rank1 as the first indicator and mAP as the second indicator. Under the same training conditions, vessels of different categories are trained as transfer datasets, and the optimal values are compared and filtered from their test results, whose corresponding vessel types are shown in Table V.

TABLE V
TYPE OF VESSELS ACHIEVING THE HIGHEST TEST
ACCURACY WITH DIFFERENT TRAINING METHODS

| Training method | The vessel type with the highest test accuracy | mAP(%) | Rank1(%) |
|---|---|---|---|
| A | / | 43.8 | 82.9 |
| AE | / | 53.2 | 91.4 |
| AC | High speed ship | 54.7 | 94.3 |
| ACE | Tanker | 58.0 | 91.4 |
| BD | Tug | 57.6 | 97.1 |
| BDF | Cargo ship | 58.2 | 100.0 |
| BDE | Fishing boat | 65.1 | 97.1 |
| BDEF | Sailboat | 63.5 | 100.0 |

From Table V, it can be seen that all six types of vessels, except passenger ships, have received the highest test accuracy under some training method, which shows that the type of vessel does not matter for the transfer dataset. And from Table I, it can be seen that the tugs, as the type with the least number of images in the dataset, are on average only 30% of the other vessels, but this does not affect their use as a transferred dataset, allowing the network to be trained to achieve the highest test accuracy.

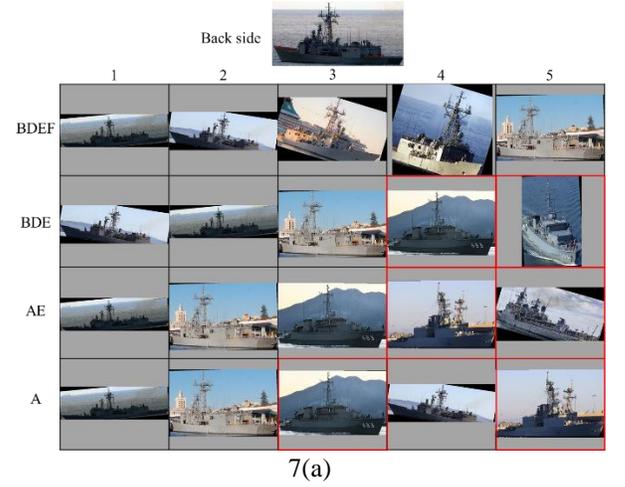

7(a)

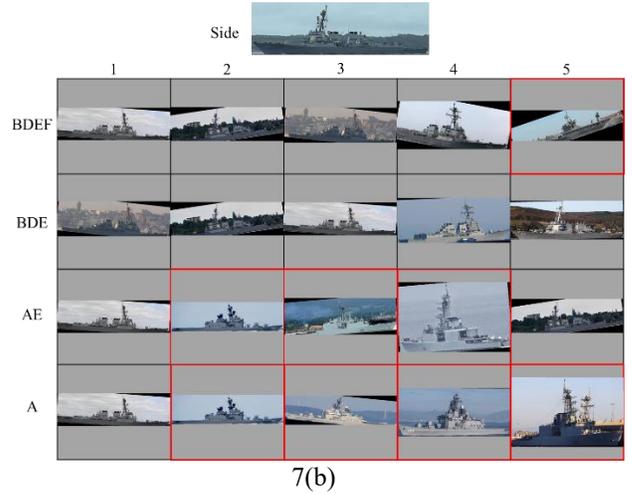

7(b)

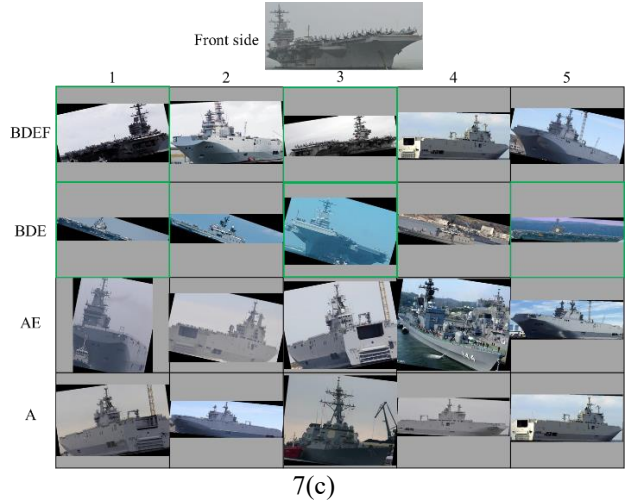

7(c)

**Fig. 7.** Effect of query set testing with different training methods, (a) Back side; (b) Side; (c) Front side, A: AlignedReid; B: Tran- AlignedReID; C: Expanded training set; D: Transferring dataset; E: Training set with SSS; F: Transferring dataset with SSS.

Figure 7 shows the results of testing the query set for three different views of the warship, with red boxes representing the incorrectly identified samples and green boxes representing the correctly identified samples, and the side with fewer samples is selected for labeling. The test results in Fig. 7(a) and Fig. 7(b) are more satisfactory. The networks trained with four different training methods have different effects, but the first graph of the search results are correct. While the test results in Figure 7(c) are average, the network trained with the transfer method performs imperfectly but the first graph of the search result is correct.

correct, and the network not trained with the transfer method fails to search for accurate results.

The proposed method in this paper is compared with several best methods, but due to the limitation of the experimental equipment, the height and width of the input image is set to 256×512, and the accuracy comparison results are shown in Table VI.

TABLE VI
RESULTS COMPARED WITH OTHER STATE-OF-THE-ART METHODS

| Method | SSS or not | mAP(%) | Rank1(%) | Rank5(%) | Rank10(%) | Weight size (MB) |
|--------|-----------|--------|----------|----------|-----------|------------------|
| Trans-Reid [28] | No | 62.5 | 88.6 | 100.0 | 100.0 | 386.62 |
| PSD-Net [29] | | 17.2 | 5.7 | 20.0 | 20.0 | 575.21 |
| AN-Net [30] | | 53.2 | 82.9 | 97.1 | 97.1 | 318.62 |
| Tran-AlignedReID | | 56.6 | 97.1 | 100 | 100 | 91.10 |
| Trans-Reid [28] | Yes | 38.3 | 68.6 | 85.7 | 94.3 | 386.62 |
| PSD-Net [29] | | 12.2 | 17.1 | 51.4 | 68.0 | 575.21 |
| AN-Net [30] | | 55.4 | 91.4 | 100.0 | 100.0 | 318.62 |
| Tran-AlignedReID | | 64.8 | 97.1 | 97.1 | 97.1 | 91.10 |

It can be seen that the Trans-Reid method achieves the highest mAP accuracy without SSS. It is proved that Transformer not only has good recognition effect in person and vehicle re-identification, but also has good recognition ability for vessels on smooth sea, but its difference with Tran-AlignedReID in terms of Rank1 accuracy is large. However, the Trans-Reid method has a general ability to identify vessels in rougher sea conditions with SSS, compared to the Tran-AlignedReID, which is very effective.

The PSD-Net method is based on the Trans-Reid method and improved for the pose of person, but it can be seen from Table VI that this method is not applicable to the vessel re-identification problem. And AN-Net is good for the re-identification of vessels, and there is some room for improvement. In addition to this, the weight size of Tran-AlignedReID is relatively small compared to other advanced methods, reflecting that it can still achieve relatively good recognition results with low network complexity.

In order to further verify the effectiveness of the algorithm, this paper is tested by the actual filming of the sea destroyer navigation video [31], whose recognition effects under different angles are shown in Fig. 8, with red boxes representing the samples with incorrect recognition and green boxes representing the samples with correct recognition, and the side with fewer samples is selected for labeling.

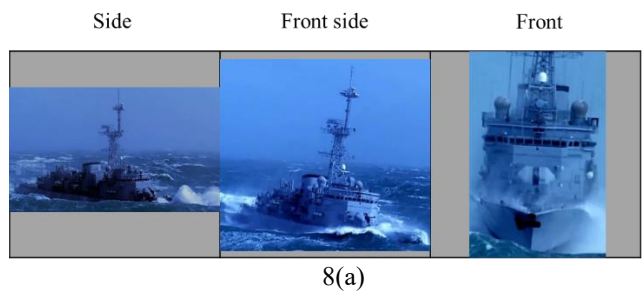

8(a)

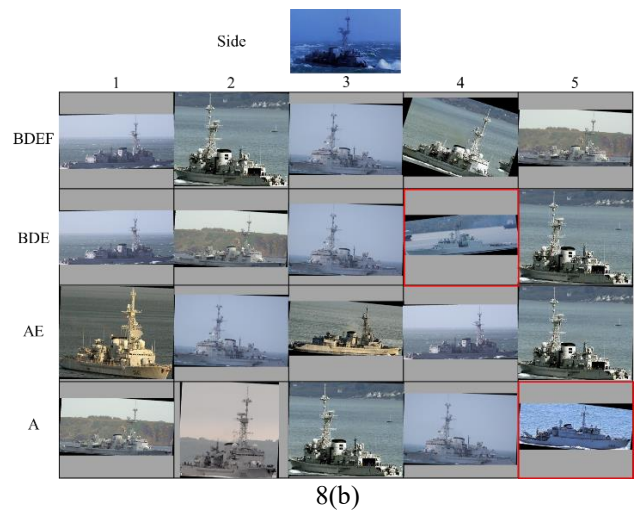

8(b)

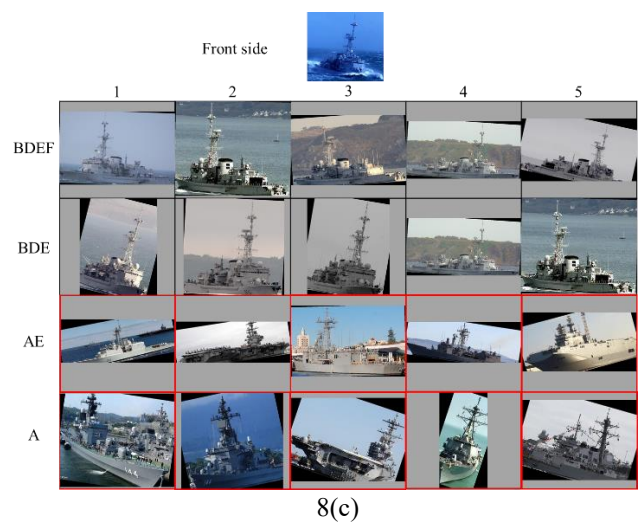

8(c)

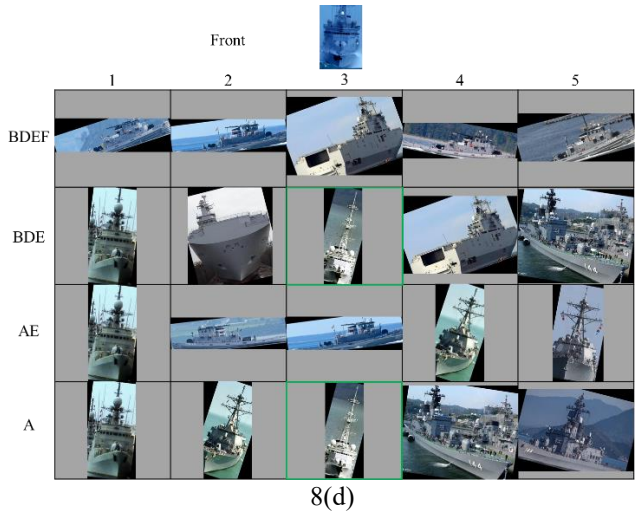

8(d)

**Fig. 8.** Effect of query set testing for video real-world experiments with different training methods, (a) Some query set images; (b) Side; (c) Front side; (d) Front, A: AlignedReid; B: Tran- AlignedReID; C: Expanded training set; D: Transferring dataset; E: Training set with SSS; F: Transferring dataset with SSS.

Some of the query set images intercepted from the video are given in Figure 8(a), including images occurring in multiple viewpoints and with different degrees of occlusion, to test the recognition performance of the algorithm for different cases of viewpoints. It is reflected in Fig. 8(b) that the algorithm before and after the improvement can recognize the vessel more accurately when the hull is obscured by waves. Comparing the recognition in Fig. 8(c), it can be seen that if the viewpoint is changed so that the stern feature of the vessel appears to be reduced and a certain angle of tilt occurs due to the high wind and waves. At this time, the algorithm before the improvement will not be able to carry out accurate recognition, while the improved algorithm can still achieve a better recognition effect after the training of SSS. However, if the vessel features are further reduced and the superstructure is missing due to the shooting problem, as shown in Figure 8(d), the algorithm cannot achieve better recognition results regardless of the improvement before and after. Therefore, compared with the vessel feature distribution, the number of features contained in the superstructure is larger than that of the hull part, and if the superstructure features are missing, it will directly lead to the degradation of the recognition ability of the algorithm.

In summary, Tran-AlignedReid improves the recognition accuracy of vessels in sea sway compared to the original algorithm, and by transferring learning with different classes of vessels, the comparison shows that the transferred dataset does not need to use a large amount of data, and the transferred dataset is not sensitive to the SSS method compared to the training set, thus proving that the Tran-AlignedReid has better robustness.

## V. Conclusion

In this paper, we propose a dynamic alignment re-identification network model fused with a transfer learning method, and perform a sea sway simulation on the dataset to simulate the swaying situation captured by on-board cameras under complex sea conditions, and repeat the experiment using different classes of vessels as the transfer dataset to explore the applicability of the transfer algorithm, and compare the effects of the sea sway simulation method on the training set and the transfer dataset, respectively. In addition, warships with high feature similarity are used as experimental objects to increase the difficulty of recognition and simulate the situation that the distant target features are not obvious due to the long sea view. The experimental results show that the improved algorithm improves on average 10.7% in mAP accuracy and 11.0% in Rank1 accuracy without sea sway simulation, while the improved algorithm improves on average 10.2% in mAP accuracy and 4.9% in Rank1 accuracy with sea sway simulation. Therefore, combined with the real-world experiments, it is shown that the transfer dynamic alignment algorithm has a higher recognition effect on the vessel re-identification task, especially in the case of vessel swaying.